# From Events to Reasoning: Hierarchical Rule Composition over Explicit States

Cheng Lyu — clyu0346@uni.sydney.edu.au

Xueqi Wang — 2317744584@qq.com

## Abstract

What a reasoning system emits is a conclusion; what it owes is a justification. Producing one requires every operation to be admissible on the facts actually available, the dependencies of intermediate results to survive to the end of the path, and the path to be revised when a premise changes. We present the Event Reasoning System (ERS), in which the reasoning state is an explicit event graph. Semantic features nominate candidate objects and rule bindings, and declared rule and task conditions decide which may execute, so candidates that violate the represented period, unit, or evidence constraints cannot enter an accepted reasoning path. Low-level rules provide reusable local transformations, and recursively composable high-level control rules organize subgoals and legal paths under the current goal, information gaps, and budget; a finite library is thereby rebound into paths never stored as complete solutions. We formalize admissible transitions, task-dependent control, compositional expressivity, dependency-preserving revision, and the gap between reachable solutions and those a budgeted controller finds. On structured-event graphs, ERS builds dependency chains of up to twelve steps, cuts irrelevant executed domain operations from 32 to 0 as lookahead grows to 16, and produces no unsupported target output across 24 negative cases; with rules frozen, it passes 12/12 focused checks comprising nine accepted derivations involving unseen input/path combinations and three justified refusals, as well as 6/6 fact-change cases. On fourteen jointly solved development tasks, the recorded dynamic subgoal configuration uses 51.6% less inference wall-clock time than single-level global control under their respective frozen settings.

## 1 Introduction

A conclusion is justified only when the operations that produced it were applicable to the facts actually available. In complex reasoning, those facts, their scope, and their dependencies must survive across multiple steps; when a premise changes, the system must also know which parts of its current solution remain valid. Natural-language reasoning can explore such problems, but the coherence of its explanation alone does not establish the admissibility of each executed dependency [1–3, 7]. The problem we address is therefore how to construct and maintain a valid reasoning process, not merely produce a plausible final answer.

We propose the Event Reasoning System (ERS). Explicit event graphs maintain object identity, time, provenance, intermediate conclusions, and their support. Low-level rules define reusable transformations with declared premises and effects. High-level control rules recursively organize those transformations into task-dependent subgoals and execution paths. The central claim is compositional: finite basic capabilities can produce new solutions through valid rebinding and dynamic hierarchical composition without storing a separate complete procedure for every task. Semantic representations help discover candidate bindings, while explicit conditions decide which operations can enter an accepted path.

ERS also makes reasoning bidirectional and revisable. High-level control directs execution; successful steps, rejected bindings, missing premises, and newly available facts feed back to reorganize the current plan. Supported results become new premises, while corrections or withdrawals trigger checks of affected downstream conclusions instead of indiscriminately discarding the entire process. Knowledge completion can extend the current premises, and historical memory can guide the search, but neither bypasses admissibility.

**Code Availability:** A runnable demo is available at https://github.com/LvCheng09/Reasoning-AI-demo-v0.1. The repository includes frontend semantic translation, event-graph construction, rule matching and execution, and output generation.

## 2 Background

### 2.1 Generative reasoning and explicit execution

Prior work on multi-step reasoning spans generative reasoning and explicit execution. CoT organizes problem-solving through intermediate reasoning steps [1], ReAct combines reasoning with environmental actions [2], and ToT further introduces evaluation and search over candidate paths [3]. The generated process may nonetheless fail to reflect the factors that determined the final answer [7].

To improve executability, PAL transforms the solution process into a program [4] and Logic-LM completes logical derivation with the aid of a symbolic solver [5, 14, 15]. A parallel line trains or applies step-level verifiers rather than accepting a final answer alone [38]. This last family shares with ERS the commitment that a step should be checked, and differs in what does the checking: a learned scorer rather than a declared condition over an explicit state. Turning open-ended input into formal input is a separate and open problem; we position ERS downstream of it rather than as a solution to it.

We introduce continuously updated explicit event states that carry object identities, applicable conditions, and evidence dependencies into the reasoning process, so that an intermediate result participates in later derivations only after a rule has checked it. Similarity-based retrieval underlies a large family of reasoning and memory systems [12, 13, 20]; we use the same machinery for candidate generation while a predicate over the state makes the final decision.

### 2.2 Planning, cognitive architectures, and the lineage of the mechanisms

Classical planning and cognitive architectures provide the foundations of structured reasoning. STRIPS achieves state transformation through conditional operators [6]; SHOP2 uses hierarchical task decomposition to organize complex goals [8]; the Options framework introduces temporal abstraction over policies [9]; and Soar supports continuous problem-solving through working memory, production rules, and dynamic subgoals [10, 11]. We build on these systems for state representation, rule execution, and hierarchical control, and treat the three as separable design concerns.

Three of the mechanisms ERS uses are older than the present setting, and stating the lineage is part of stating the contribution. A rule with a required pattern, a produced effect, and an execution operator is an operator schema over a typed state [29], and a precondition paired with an effect is a Hoare-style specification of a partial operation [28]. The qualification problem motivates recording known defeating exceptions within the rule applicability conditions [30]. Revision directed by recorded justifications, and the distinction between *no longer supported* and *disproved*, belong to truth maintenance [35, 36]; the status vocabulary of absent, withdrawn, and contradicted draws on closed-world and non-monotonic accounts of negation [31, 32, 33] and on theory change [34], and validity intervals over facts are the interval algebra of [37].

What ERS adds to these foundations is not the mechanisms themselves but their integration: an explicit interface from graded semantic candidate discovery to rule-level admissibility, bidirectional execution-to-control feedback, and versioned capability editing that keeps knowledge, execution rules, and control strategies independently maintainable.

### 2.3 Relation to world models and generative architectures

World models aim to let an agent predict possible futures by learning internal representations of its environment and planning against them [22, 23]. Dreamer-style agents predict the outcome of actions and optimize policies by imagining trajectories [25], and JEPA-style architectures learn semantic representation by prediction in feature space rather than by reconstructing inputs [24, 26]; hierarchical variants extend this to planning at several temporal scales [27]. ERS shares with this literature the view that a complex world can be characterized by the features of things and their relations, and differs in how that representation is used: learned features provide semantic representations and nominate candidate bindings, while explicit rule and task conditions determine which operations may change the state.

We intend the comparison in Table 1 as architectural rather than empirical: each property follows from the design assumptions of the family in question, and the table delimits the intended division of labor rather than reporting measured performance.

Three differences bear on the claims we make here. Prediction quality and logical validity are different properties: a latent model can produce a statistically plausible next state without any component whose function is to check whether a particular step was admissible. Continuous similarity and discrete admissibility serve different roles: this literature uses latent proximity to measure, while ERS uses it only to propose candidates and reserves the decision for explicit conditions. Correctness of an answer and validity of the process are separate: a scoring rule based on resemblance to a reference trajectory penalizes an admissible path that differs from the demonstration, of which Section 6.5 reports an instance.

**Table 1.** Positioning of ERS against three families of methods, by stated design assumption rather than by measured performance.

| Property | CoT / ReAct / ToT [1–3] | Program and solver hybrids [4, 5] | World models [22–27] | ERS |
|---|---|---|---|---|
| Typical intermediate representation | language and/or tool traces | program or symbolic form | learned latent states | explicit event graph |
| Local rule applicability | may use external checks | defined by program/solver semantics | model-dependent; not always explicit | explicit predicate over facts and task constraints |
| Evidence dependencies | method-dependent | program/solver dependent | often implicit in latent states | explicit in accepted event supports |
| Revision after fact change | varies by implementation | recompute or maintain proof state | model/system dependent | dependency-scoped revalidation |
| Failure diagnosis | varies by implementation | program/solver traces | model/system dependent | explicit checks and traces, with some external diagnoses |
| Capability extension | prompting and/or tools | program/solver and module changes | learned or hybrid updates | editing knowledge, execution, and control rules |
| Unsupported conclusion handling | system dependent | solver/system dependent | application dependent | unresolved state permitted |

## 3 Model Architecture

We model reasoning as the continual construction of an executable path over an evolving event state. The architecture has four interacting elements: semantic access, which produces candidate concepts and bindings; explicit events, which maintain objects, relations, evidence, and accepted results; low-level execution rules, which implement local state changes; and high-level control rules, which organize goals and choose how the local rules are connected. Knowledge and memory provide complementary support (Figure 1).

The principal interaction is bidirectional. Control selects a subgoal and a legal candidate operation using the current state; execution then changes that state—or records a failure without admitting an unsupported result—and returns information that changes the next control decision. We therefore form the hierarchy from the task and revise it by execution, rather than prescribing a fixed sequence.

A separate capability-management process may change the knowledge and rule libraries between reasoning episodes. Sections 3.5–3.6 detail the difference between control adaptation and inference-time knowledge completion.

For consistency, $t$ denotes an inference-time state or control step, $u$ a longer-term capability-library version, and $q$ the current task. We write $R_{exec}$ and $R_{ctrl}$ for execution and control rules, and $K$ and $M$ for accessible knowledge and historical memory. A task budget $B_q$ constrains the resources consumed by reasoning. All rule effects remain subject to explicit acceptance checks.

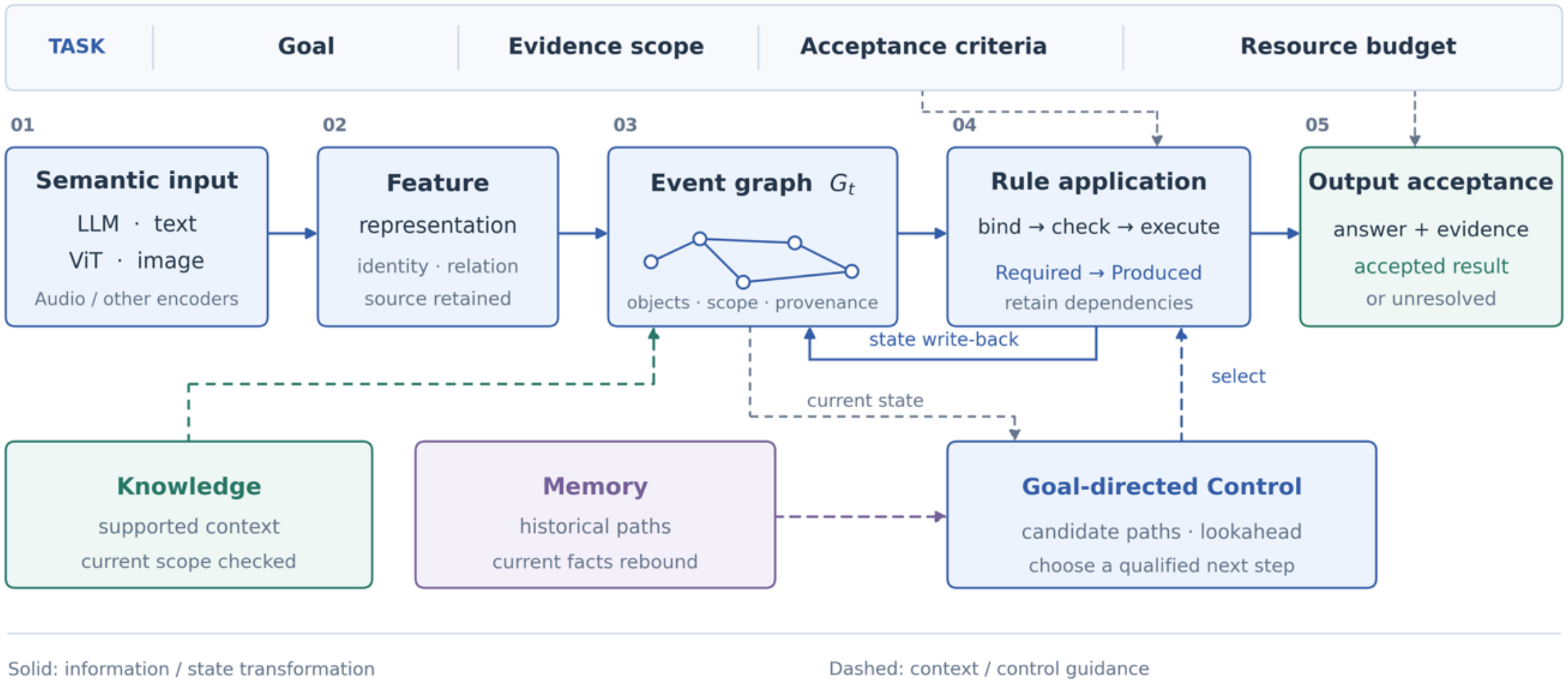

**Figure 1. ERS** reasoning architecture. Rule application denotes low-level execution, while goal-directed control organizes goals and strategies through high-level rules. Evidence scope and acceptance conditions are task constraints. Modality-specific encoders produce features that are organized into an event graph preserving object identity, evidence scope, and provenance. Control selects rules from the current state and objective; execution checks bindings and applicability, then writes back results and dependencies. Knowledge provides evidence-scoped information for admissible completion, while memory guides path selection through rebinding to current facts. Solid arrows denote information or state transformations, and dashed arrows denote contextual or control guidance. The diagram shows general input interfaces; the reported experiments evaluate the reasoning backend using structured event graphs.

### 3.1 Feature-based semantic representation

We start from the assumption that objects can be described through features and relations. Language, vision, and other inputs may express overlapping properties while also contributing modality-specific information. We keep continuous semantic representations as a graded interface to these descriptions, while exact quantities and object identities remain explicit for rule execution.

A semantic unit is

$$v_i = \langle id_i,\ z_i^c,\ z_i^r,\ z_i^n,\ a_i,\ \sigma_i \rangle. \tag{1}$$

Here $z_i^c, z_i^r, z_i^n$ encode concept, relation, and attribute features; $a_i$ contains values, units, and other payloads that require exact operations; and $\sigma_i$ contains evidence, identity, and temporal qualifiers. We keep the exact payload separate from the graded representation, so that a high similarity score is never itself authorization to use a value.

For nonzero vectors in the same feature channel, candidate matching can use

$$s_{i,j}^k = \frac{\langle z_i^k, z_j^k \rangle}{\|z_i^k\|_2 \|z_j^k\|_2}, \qquad k \in \{c, r, n\}. \tag{2}$$

Similarity generates a candidate set. Its threshold changes how many candidates are examined, but a candidate enters an executed path only after the discrete checks defined in Section 3.4. We call this the nomination–admission separation; it lets semantic access and rule applicability be extended independently.

For modality $m$, a general encoding interface is

$$F_m = E_m(x_m), \qquad Z_m = P_m(F_m), \tag{3}$$

where $E_m$ is a modality encoder and $P_m$ maps relevant features into a comparable space. Alignment preserves source identities and modality-specific information rather than assuming that all modalities yield identical features [12, 13, 19]. The mapped information is organized into addressable events by resolving candidate object identities, attaching source references, preserving observation times and effective periods, and recording relations only when the event's evidence contract permits them. An unresolved match remains a candidate or hypothesis; semantic proximity does not turn it into an accepted fact. These organization and qualification steps form the interface between flexible semantic input and executable rule binding. Our experiments begin from structured events and therefore study the reasoning backend rather than the quality of cross-modal grounding.

### 3.2 Explicit events, evidence, and changing state

We organize each local problem as an event: the objects, semantic units, relations, observations, intermediate results, goals, and evidence it concerns. We link events into a dependency graph: observations and given conditions support rule executions; accepted executions produce intermediate facts; and those facts may support later subgoals. Several premises may jointly support a conclusion, and different valid paths may reach the same goal. Every accepted rule application records its actual bindings, source rule, output, and supporting facts. The resulting dependency structure distinguishes joint premises from alternative support: all premises within one justification must hold together, while a conclusion with multiple independent justifications may remain supported through any one valid alternative. Execution history, dependency structure, and candidate branches are consequently different views of the same episode; a chronological trace alone need not capture every branch.

Identity, event boundaries, and temporal scope. A fact carries the identity of the object it concerns, a source reference, an authorized evidence scope, and a temporal record. Information may enter another event only through an explicit inter-event relation or a supported, scope-compatible evidence link; belonging to the same semantic category is insufficient for sharing it. For dynamic facts, the qualifier $\sigma_i$ distinguishes recording time $t_i^{add}$, the observation or effective interval $I_i^{eff}$, and, if applicable, invalidation time $t_i^{inv}$; it also retains status and invalidation reason. These times serve different purposes: importing an FY2024 record in FY2025 does not make its revenue applicable to FY2025. Invalidated and superseded entries retain their original sources, timestamps, and dependencies for reconstruction, while the active event graph admits only currently supported facts applicable to the task. We maintain inference dependencies and the chronology of revisions separately and consult them together.

Support and non-monotonic revision. We treat an accepted derived fact $x$ as carrying one or more recorded supporting justifications, each identifying the premises and rule application that produced it. When a premise is replaced or withdrawn, its dependent conclusions are reconsidered. A conclusion remains accepted if a valid independent support survives; otherwise its status becomes pending verification or invalid as required by the task's acceptance policy. Withdrawal is not automatically negation: *absent*, *withdrawn*, and *contradicted* support represent different conditions. This is a dependency-aware, non-monotonic form of maintenance related to truth-maintenance systems [35, 36], embedded here in the broader event-and-control architecture.

The dependency record must avoid self-justifying circular support: a conclusion cannot remain accepted solely because a cycle of conclusions refers back to itself. We therefore require an independently grounded support base for accepted justification chains, and Section 4.2 states the associated locality property. Changes to the rule library are handled on the capability timescale described in Section 3.6; their affected derivations are checked through the same dependency mechanism.

### 3.3 Tasks and reasoning states

We specify a task by a goal, the constraints governing legitimate evidence and operations, and the available resources:

$$q = \langle g_q,\ C_q,\ B_q \rangle. \tag{4}$$

The goal $g_q$ defines a satisfying result; $C_q$ specifies allowed objects, time periods, units, conventions, and evidence scopes; and $B_q$ is a resource budget. We treat the task constraints as operative conditions rather than incidental prose: a calculation using the wrong reporting period may be numerically correct and still fail the task.

The reasoning state is

$$S_t = \langle G_t,\ q,\ H_t \rangle, \qquad G_t = \langle V_t,\ L_t,\ F_t,\ \Sigma_t \rangle. \tag{5}$$

Here $V_t$ is the collection of semantic units, $L_t$ contains typed relations and dependencies, $F_t$ contains the active focus and subgoals, $\Sigma_t$ contains qualifiers and statuses, and $H_t$ records accepted executions and relevant failed attempts. If a user changes task constraints, the task is revised and dependent conclusions are reconsidered; the notation $q$ denotes the currently active task version. We keep the subgoals in two places for two purposes: $F_t$ records the active focus and subgoals as references visible in the event state, while the control state $Z_t$ defined in Section 3.5 is authoritative for their ordering, prioritization, and resource allocation. After a control update, the event graph's focus references are synchronized with the current control organization, so the two representations cannot silently disagree.

### 3.4 Low-level rules: reusable and admissible local capabilities

A low-level rule declares a local capability, such as arithmetic, unit conversion, comparison, or relational inference. It is reusable because variables are rebound to current facts; the rule is not an entire stored solution trajectory. We write

$$r = \langle Req_r,\ Prod_r,\ C_r,\ \Phi_r \rangle, \tag{6}$$

where $Req_r$ is the required premise pattern, $Prod_r$ the expected effect, $C_r$ the applicability conditions (including task-relevant evidence restrictions), and $\Phi_r$ the execution operator. For binding $b$, admissibility is

$$Adm(r, b; S_t) = \mathbf{1}[Req_r(G_t, b)\ \wedge\ C_r(G_t, b, q)]\,. \tag{7}$$

A semantic match only nominates a candidate. The first conjunct requires grounded premises; the second checks object identity, units, time, exceptions, and the task's allowed evidence. Repeated occurrences of a variable must bind consistently. An incorrect semantic interpretation that is already encoded as an apparently valid fact is not detectable by this predicate without conflicting or external evidence; we treat this as a property of the grounding boundary, and it is the reason we keep the checks explicit rather than implicit in a similarity score.

For an admitted binding, $\Phi_{r,b}$ proposes the next state, which is accepted only if its declared effect holds:

$$S_{t+1} = \Phi_{r,b}(S_t), \qquad Adm(r, b; S_t) = 1, \quad Prod_r(S_t, b, S_{t+1}) = 1. \tag{8}$$

If the premises are absent or the effect is not satisfied, the operation produces no *accepted domain-state update*. The rejected attempt, its bindings, the violated condition or effect, and the resources consumed are nevertheless recorded in $H_t$ and returned as control feedback. Failure therefore leaves the accepted facts unchanged while extending the execution trace, and it cannot be treated as a successful premise. Successful execution writes back its result and the provenance of every dependency, making it available as a legitimate premise for subsequent operations. We keep the local execution rule and the higher-level decision to invoke it as distinct mechanisms.

### 3.5 Dynamic hierarchical composition and bidirectional feedback

Two levels, recursively organized. We use two rule levels rather than one, because what a state permits and which subgoal to pursue next are decided from different information. Low-level rules determine which operation is legal; high-level control rules determine which subgoal should be pursued, which local operations should be connected, and how the path should change when information or execution outcomes change. A control rule may invoke a more specific control rule, creating as many levels as the task and the resource budget permit. The number of control levels is not predefined by the rule library, and need not equal either the eventual execution-path length or the lookahead depth.

The rule library is partitioned by function:

$$R = R_{exec} \cup R_{ctrl}. \tag{9a}$$

The active control state records the selected subgoal, dependency ordering, candidate organization, and remaining resources:

$$Z_t = \langle \zeta_t,\ \preceq_t,\ \mathrm{Cand}_t,\ \mathrm{Bud}_t \rangle. \tag{9b}$$

A high-level rule $c = \langle Req^c, Prod^c, \Psi_c \rangle$ specifies when an organizational transformation applies and what control state it produces:

$$Z_t^+ = \Psi_{c_t}(S_t, Z_t; q), \qquad Req_{c_t}^c(S_t, Z_t, q) = 1, \quad Prod_{c_t}^c(S_t, Z_t, Z_t^+) = 1. \tag{9c}$$

Top-down guidance. Given the active goal and graph, control decomposes a target into dependent subgoals, selects eligible strategies, and allocates exploration. For example, comparing two financial indicators may require harmonizing conventions, computing each indicator, and finally comparing the results. A subgoal can be divided further until the corresponding operations are executable by low-level rules. We *construct* this decomposition rather than retrieving a complete memorized answer. The control hierarchy describes which strategies and subgoals are currently organized; the event dependency graph describes which facts and operations support the accepted conclusions; and $H_t$ orders actual attempts in time. A branched solution can therefore be represented without mistaking a linear execution log for the complete reasoning structure.

Bottom-up and state-change feedback. A low-level attempt reports success, unmet premises, failed applicability, inconsistent effects, newly supported facts, unresolved dependencies, and resources consumed. Control also responds to facts added or withdrawn outside the current attempt, to supported knowledge completion, and to changes in the task's constraints. Let $\Delta G_t$ and $\Delta q_t$ denote these optional event and task changes. The feedback and control update are

$$f_t = Feedback(S_t, Z_t^+, \mathrm{attempt}_t, \Delta G_t, \Delta q_t), \qquad Z_{t+1} = UpdateCtrl(Z_t^+, S_{t+1}^*, f_t; q_{t+1}),$$

where $S_{t+1}^*$ contains only accepted domain changes, and $q_{t+1}$ is the active task after any authorized revision. An unsuccessful attempt can leave the accepted graph unchanged while adding a failure record to $H_t$. Likewise, an external correction can trigger this update without a low-level operation having just succeeded. The controller may reorder subgoals, invalidate a now-inapplicable branch, preserve independently supported results, activate an alternative legal continuation, or seek missing premises. Control directs execution, and verified execution or external state changes redirect control; this is the mechanism by which the reasoning hierarchy is rebuilt around the current problem rather than fixed in advance.

Local control and retained state. For an active subgoal, a control strategy may evaluate admissible candidates against the local dependencies and retain accepted intermediate results instead of restarting the full goal search after each attempt. A failed low-level operation first changes the candidates or continuation within the affected subgoal; if its required premises cannot be restored or the current strategy is no longer applicable, feedback can propagate to the enclosing control decision and reorganize the higher-level subgoals. Retained results remain available only while their recorded supports satisfy the active task constraints. This local-first organization does not prescribe a fixed number of control levels or guarantee a lower cost on every task.

Figure 2 illustrates the top-down guidance and bottom-up feedback described above, including failed applicability checks, accepted event-state updates, and task or evidence changes that prompt a revised control decision.

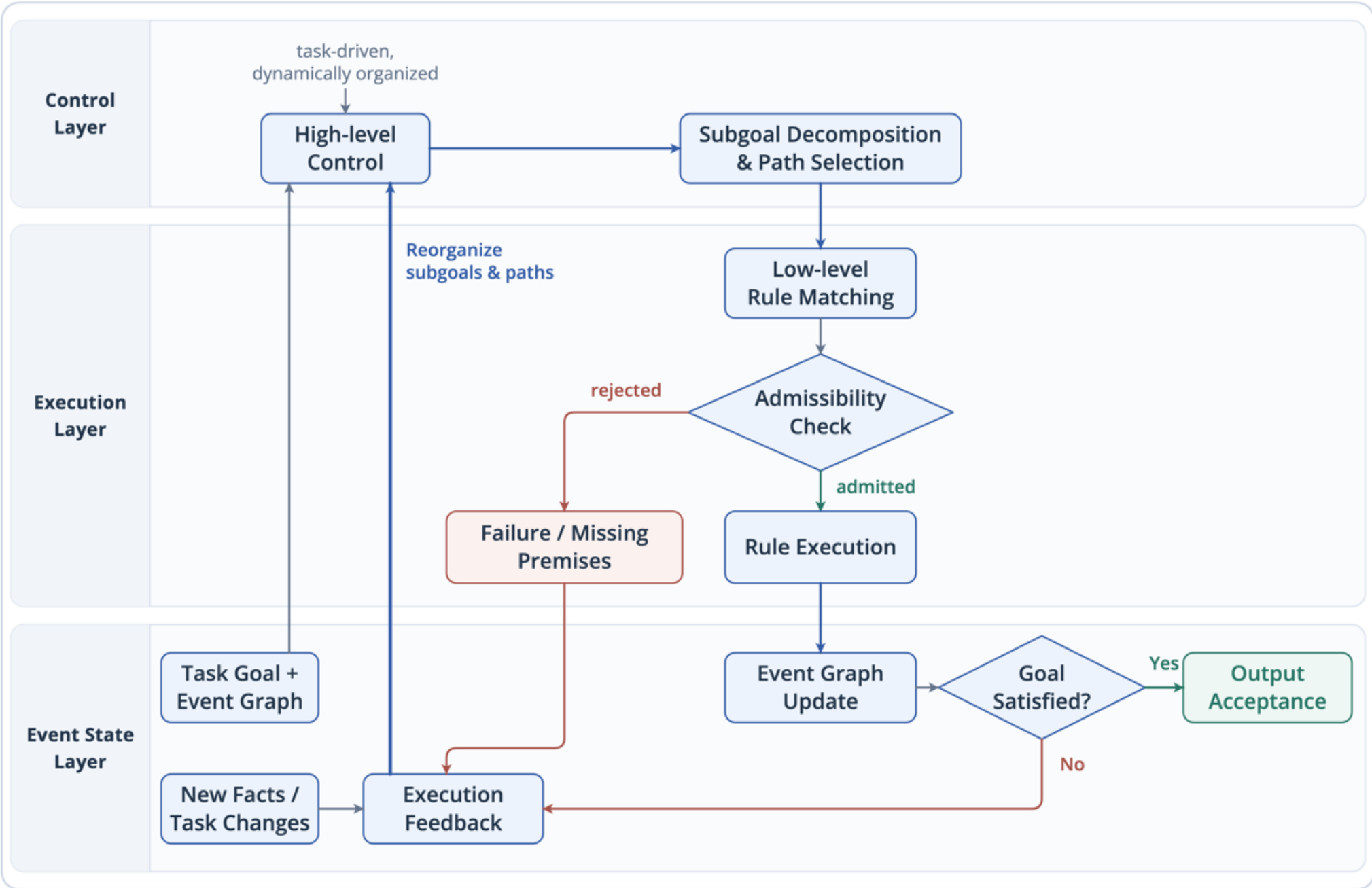

**Figure 2.** Dynamic hierarchical control and bidirectional feedback in ERS. Goal-directed control organizes subgoals and selects candidate paths; low-level rule matching and admissibility checks govern accepted state changes. Rejected attempts, missing premises, and changes to facts or task constraints return feedback that can reorganize the active subgoals and paths. Output acceptance remains subject to the task and evidence conditions in (13).

Candidate discovery and selection. The current control state and required patterns guide retrieval:

$$D_t = Discover(S_t, Z_t^+; R_{exec}), \tag{10}$$

$$A_t = \{(r,b) \in D_t : \; Adm(r,b;S_t) = 1\}. \tag{11}$$

If $A_t \neq \varnothing$, the controller chooses an admitted rule–binding pair under goal progress, dependencies, available knowledge, memory guidance, and lookahead depth $h$:

$$(r_t, b_t) \in \arg\max_{(r,b)\in A_t} V_h(S_t, r, b; q, K, M, Z_t^+), \qquad S_{t+1} = \Phi_{r_t,b_t}(S_t). \tag{12}$$

The state transition in (12) is understood to require the effect check in (8); a rejected candidate does not modify the accepted graph. The evaluation function ranks admissible candidates and does not override their admissibility. Control can choose additional search when no candidate is admitted, including knowledge completion, strategy switching, or returning an unresolved result.

Adaptive depth and stopping. The hierarchy has no universally fixed depth. Each active expansion must have an applicable control rule, a well-defined subgoal or dependency change, and a stopping condition. Exact repeated control configurations can be detected from the goal, relevant state, rule-library version, and unresolved premises. With a finite budget that charges a positive amount for each expansion, any *budgeted inference episode* terminates after finitely many charged expansions; we do not assert unconditional termination for an unrestricted recursive rule system. Goal satisfaction, a directly executable subgoal, exhaustion of valid alternatives, or budget exhaustion can also stop a branch. These controls prevent unproductive recursion while retaining task-dependent depth.

### 3.6 Knowledge completion and experience-based guidance

We separate two inference-time roles: knowledge supplies missing premises, and memory guides path search. Knowledge completion may add a supported property or relation to the current event graph, whereas historical memory proposes previously useful rule paths or candidate priorities. A remembered conclusion does not become evidence for the present task merely because its earlier execution succeeded.

A supported completion is written $G_t \mapsto G_t \oplus \Delta_K$, where $\Delta_K$ identifies the referenced objects, supporting evidence, and permitted event scope. Definitions can justify semantic or relational completion but do not authorize an unobserved measurement. After each accepted increment, the system re-evaluates previously blocked Required conditions, and a newly derived conclusion may expose another missing premise or support a subsequent completion round. Each round has its own evidence checks, and recursion remains bounded by the task's resource budget and control termination rules.

Memory retains the initial event conditions, execution path, outcome, and constraints that mattered to success or failure. Its reuse requires fresh object binding, task-specific applicability checking, and independent evidence acceptance. It can propose a different branch or reorder candidates but cannot substitute an old value for a currently required fact. In the experiments we report here, the persistent knowledge and rule collections are held fixed; forming a new path from existing rules is an inference-time act, independent of any later modification to the rule library.

### 3.7 Output acceptance

Stopping an inference episode does not imply that the goal has been achieved. ERS accepts a conclusion only when the goal, task constraints, and actual evidential support are satisfied:

$$Accept_q(S_T, H_T) = \big[g_q(G_T) \ \wedge \ C_q(G_T, H_T) \ \wedge \ Evidence_q(G_T, H_T)\big]. \tag{13}$$

The evidence condition requires that the invoked facts and justifications have valid identities, applicable scopes, current support, and admissible rule histories. The system distinguishes accepted success from unresolved premises, exhausted search, a rejected operation, and unsupported output. Alternative paths need not be identical to a stored demonstration; they are acceptable if they satisfy the same goal and validity requirements. We keep task success, path validity, and conformance to a particular reference mechanism as three separate quantities.

## 4 Finite Rules and Open Compositional Reasoning

ERS is built around a distinction among basic capability, path composition, and task-dependent organization. The rules stored by the system specify reusable local transformations and control strategies; the paths produced during reasoning depend on how those rules are rebound and connected to solve the present task. We formalize below the difference between what a rule library can express, what it can legally reach, and what a controller can actually discover under a budget.

### 4.1 Finite descriptions, extensible instances and paths

Modeling premise. We assume that, within a specified domain, knowledge scope, and representation accuracy, a finite collection of reusable local relations may suffice to generate many distinct problem-solving processes. The number of complete tasks need not equal the number of stored rules. A social or institutional task uses conventions that may be revised; a natural-law task relies on objective constraints; both can be represented through rule patterns whose variables bind to particular objects, circumstances, and goals. We state the premise as conditional on rule coverage and legal applicability.

For a valid path $\pi = \langle (r_1, b_1), \ldots, (r_L, b_L) \rangle$, define its state transformation and the reachable set

$$\Phi_\pi = \Phi_{r_L,b_L} \circ \cdots \circ \Phi_{r_1,b_1}, \qquad Reach_R(S_0) = \{\Phi_\pi(S_0) : \pi \text{ is a valid path}\}.$$

We call a *valid path* any sequence whose bindings satisfy (7), whose effects satisfy (8), and whose intermediate support remains admissible. Let $\mathcal{G}_q$ denote the goal-satisfying states. Then $Reach_R(S_0) \cap \mathcal{G}_q \neq \varnothing$ describes a reachable target under the current facts and rules. Reachability does not imply that a finite-budget controller will find the path.

Rebinding. The same arithmetic or relational rule can apply to new objects, values, or periods without storing another copy of the rule. The existing basic property may be stated as follows.

**Proposition 1 (Conditional unbounded instance expressivity of finite rules).** Suppose a fixed finite rule collection can legally apply a local pattern to states $S_n$ for every $n \in \mathbb{N}$, and the resulting states $S'_n$ are mutually distinguishable. The collection then describes infinitely many valid transformation instances through binding.

*Proof.* Each $n$ gives an admitted instance of the same stored pattern, and the outcomes are distinct by assumption. Thus infinitely many instances are described using the fixed rule collection. ■

The proposition concerns instances rather than tasks: what it makes explicit is the separation between the number of rule descriptions and the number of legal bindings.

Composition adds a second dimension. Let $\mathcal{B}$ be the binding domain and $\mathcal{L}_{ctrl}$ the control-permitted sequences of low-level rules. The legal path space is

$$\begin{aligned} P_{ctrl,\mathcal{B}}(S_0) = \bigcup_{L\geq 0} \Big\{ & \langle (r_1, b_1), \ldots, (r_L, b_L) \rangle : \\ & (r_1, \ldots, r_L) \in \mathcal{L}_{ctrl}, \\ & b_t \in \mathcal{B} \quad (1 \leq t \leq L), \\ & Adm(r_t, b_t; S_{t-1}) = 1 \quad (1 \leq t \leq L), \\ & Prod_{r_t}(S_{t-1}, b_t, S_t) = 1 \quad (1 \leq t \leq L) \Big\}. \end{aligned} \tag{14}$$

If legal rule composition permits arbitrarily long finite paths—for example, when a repeatable local transformation has a supported successor state at each finite iteration—then a fixed finite description can generate paths of unbounded length. Conversely, a finite acyclic dependency structure with bounded bindings may admit only finitely many paths. The expressive conclusion therefore follows from composability and admissibility, not from finiteness of the rule library alone. A separate line of work measures a related but distinct quantity, namely whether a learner transfers to combinations of primitives it has seen separately [16, 17, 18]; that question is a property of a learner, whereas legal reachability is a property of the path space a fixed library admits.

Let $\Pi_{\text{stored}}$ denote complete instantiated rule–binding paths recorded before the current evaluation, and let $\mathcal{L}_{\text{seen}}$ denote complete ordered rule sequences previously instantiated in stored control or execution traces. Unlike the historical sets, $\mathcal{L}_{\text{ctrl}}$ specifies the rule sequences permitted by the active control organization, regardless of whether they have been previously executed.

For a task input $x$ with initial state $S_0(x)$, a valid path $\pi \in P_{\text{ctrl},\mathcal{B}}(S_0(x))$ is an unseen complete path when $\pi \notin \Pi_{\text{stored}}$. It further exhibits unseen rule-sequence composition when $\text{rules}(\pi) \notin \mathcal{L}_{\text{seen}}$.

New bindings of an existing sequence and previously unused sequences are thus distinct forms of composition; neither requires storing a separate complete solution for every task. This definition concerns the system's stored and executed paths, rather than unseen samples in a model-training dataset.

At a branching point, a high-level strategy may connect the same local prefix to different admissible continuations depending on the goal. Figure 3 illustrates two paths formed from one small rule collection.

More generally, the hierarchy composes rules *and* strategies: a high-level strategy may call more specialized strategies before committing to executable steps, and no separate complete path has to be stored for each combination.

The path $\pi$ denotes one valid executable ordering. Where the problem has converging dependencies or legal alternatives, a branching dependency structure can support multiple such orderings and high-level decompositions.

The same reusable library therefore supports both newly reachable tasks and better organized solutions to already reachable tasks—for example, by replacing an unnecessary branch with a shorter admissible continuation. Path quality, reachability, and execution cost remain distinct concepts. This establishes the central organizational idea: finite capabilities are the elements, and their goal-conditioned compositions are the realized reasoning structures.

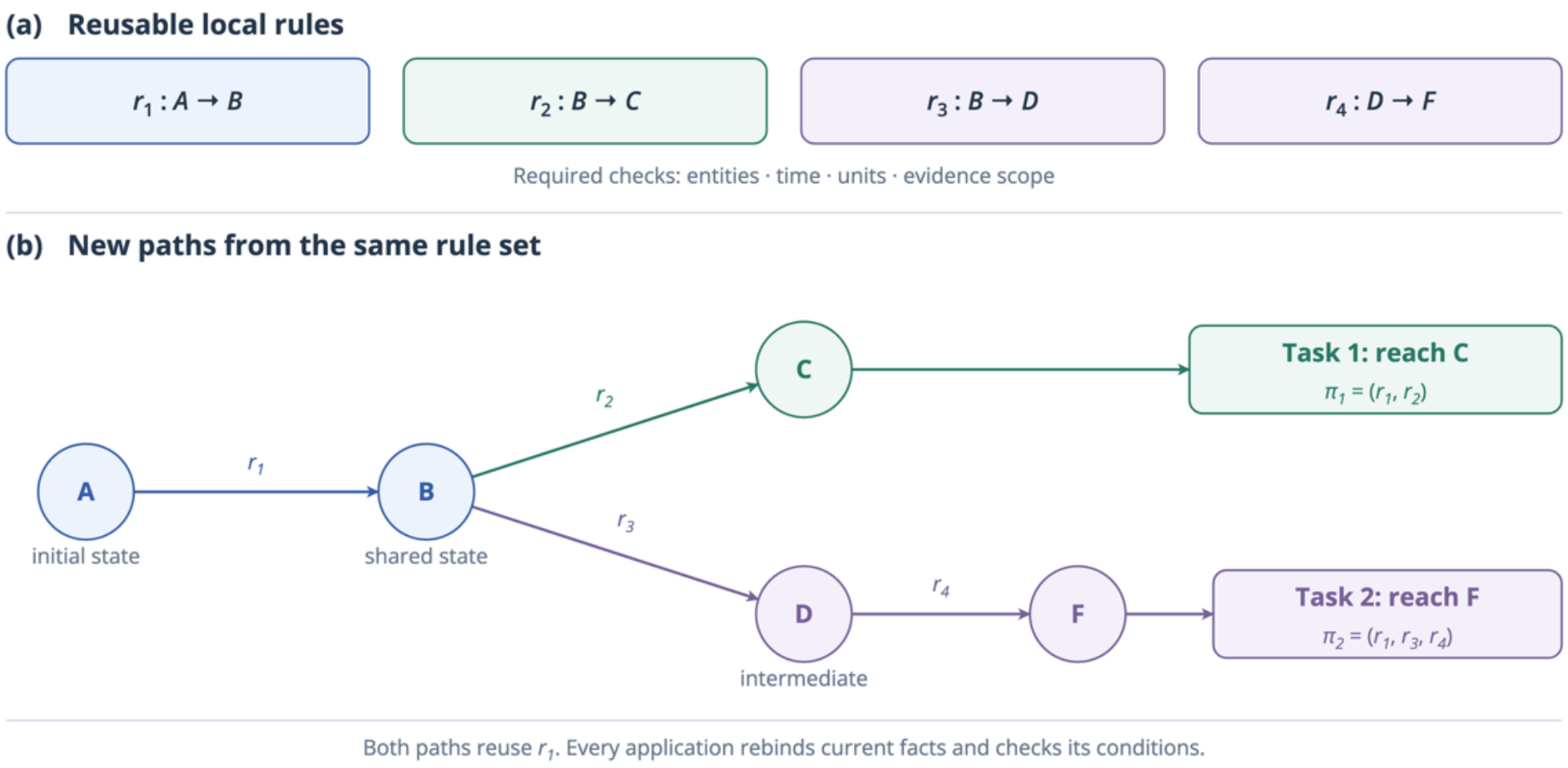

**Figure 3.** Reusable local rules and goal-conditioned path composition. Rules $r_1$–$r_4$ are defined once; the same initial transition $r_1$ participates in a path reaching $C$ and an alternative path reaching $F$. Each application rebinds the current objects and is subject to the declared identity, time, unit, and evidence conditions.

### 4.2 Accepted paths, dependencies, and local revision

The ability to combine rules is useful only if the composed path preserves its declared conditions. A locally correct arithmetic result can still be an invalid step when it depends on an incompatible object, period, unit, or evidence source. We check applicability before execution and record the factual and rule dependencies of accepted outputs, which connects local operations into an inspectable derivation structure.

Let $\mathcal{I}$ denote a collection of declared state invariants such as type compatibility, bound-object identity, permitted evidence scope, and valid support for active conclusions.

**Proposition 2 (Conditional path consistency).** If $\mathcal{I}(S_0)$ holds and every *accepted* transition preserves $\mathcal{I}$ whenever it is admissible, then $\mathcal{I}$ holds after any finite sequence of accepted transitions:

$$\mathcal{I}(S_0) \;\wedge\; \bigwedge_{t=0}^{L-1} [\mathcal{I}(S_t) \Rightarrow \mathcal{I}(S_{t+1})] \;\Rightarrow\; \mathcal{I}(S_L). \tag{15}$$

*Proof.* The base state satisfies $\mathcal{I}$. Each admitted and accepted transition preserves $\mathcal{I}$, so induction on the finite path length gives the result.

The property guarantees consistency with respect to the declared conditions and states precisely how local checks compose across a long execution path; it does not by itself certify the factual truth of an input or the correctness of a misspecified domain rule.

Dependency-scoped revision is a separate property. For a derived conclusion $x$, let $Just(x)$ contain its alternative justifications, where each justification is the collection of jointly required premises together with the rule and conditions that authorize the derivation. Then $x$ remains supported exactly when at least one such justification is valid: $Supported(x) \Leftrightarrow \exists J \in Just(x)$ such that all premises in $J$ have independent valid support and the producing rule's declared checks hold. Let $Dep^*(f)$ be the transitive set of conclusions whose recorded justifications refer to fact $f$, directly or through intermediate results. Assume that only $f$ changes, that the task and applicable rule versions remain fixed, and that accepted justification chains have an independently grounded base. Then conclusions outside $Dep^*(f)$ retain their existing support. A conclusion inside the dependency closure remains accepted if at least one unaffected justification remains valid. Otherwise, affected justifications are revalidated against the updated premises; this may produce a revised conclusion, while the original conclusion is marked unresolved or unsupported if its required support no longer holds. The *candidate scope* of revalidation is the dependency closure $Dep^*(f)$, not every conclusion in the graph.

We separate the two because they answer different questions. Path consistency explains why a successful derivation remains valid under its conditions; dependency-scoped revision explains what happens when a former condition ceases to hold. A withdrawn premise is not thereby proved false, and a result with an unaffected independent justification need not be discarded. The underlying justification principle is related to TMS/ATMS [35, 36]; we use it as part of continuously evolving event states and goal-directed control. If task constraints or rule versions change globally, the affected scope may be broader than $Dep^*(f)$ and must be recomputed accordingly.

Figure 4 illustrates how a change to a shared premise invalidates dependent support paths without necessarily removing a conclusion that retains an independent valid justification. Withdrawal and contradiction remain distinct evidence statuses; the diagram depicts the propagation of lost support, not an automatic inference that the withdrawn premise is false.

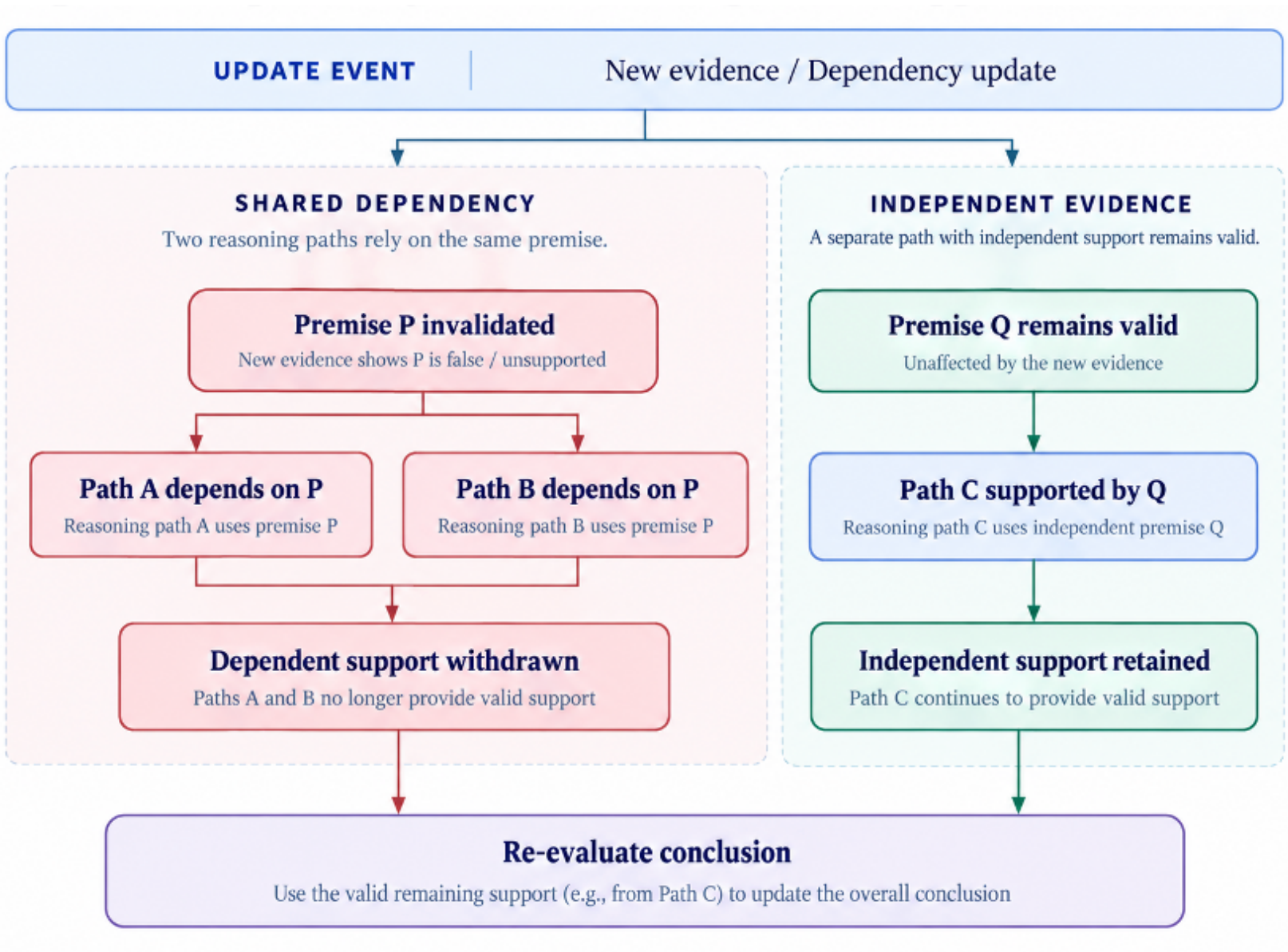

**Figure 4.** Dependency-scoped revision with alternative supports. Paths A and B require premise $P$ and lose that support if $P$ is invalidated or withdrawn. Path C is grounded in an unaffected premise $Q$ and remains a valid independent justification. The shared conclusion is re-evaluated using its remaining supported paths, rather than being discarded solely because one premise changed.

### 4.3 Reachability, control, and the sources of failure

For task distribution $\mathcal{D}$ and controller $\mathcal{C}$, we define budgeted *potential coverage* and *realized control capability*:

$$Cov_B = \Pr_{q\sim\mathcal{D}} \left[\exists \pi : \ Cost(\pi) \le B \ \wedge \ Accept_q(S_T^\pi, H_T^\pi) = 1\right], \tag{16a}$$

$$Cap_B(\mathcal{C}) = \Pr_{q\sim\mathcal{D}} \left[\mathcal{C} \text{ returns an accepted valid path using total cost } \le B\right]. \tag{16b}$$

Here $S_T^\pi, H_T^\pi$ denote the final state and history obtained by executing path $\pi$; $Cost(\pi)$ is its domain execution cost. Under a common task and budget unit and the assumption that a controller's total search-and-execution cost is at least that of the path it returns,

$$Cap_B(\mathcal{C}) \le Cov_B. \tag{17}$$

This is a structural upper bound rather than an estimate computed by ERS, and it distinguishes a legal path's existence from a controller's ability to find it. The distinction matters because additional search, evidence, or rules widen what becomes reachable and discoverable. The three levels are therefore distinct: expressible compositions, reachable goal states, and successfully discovered and executed solutions within budget.

Failure location and causal attribution. We make observable failures inspectable through explicit execution records: a retrieved rule can fail a requirement check; a binding can violate the task's period; an execution can fail its declared effect; a controller can exhaust its allocated budget. The *cause* of a symptom may nonetheless require additional evidence. An erroneous semantic encoding can appear locally identical to a rule-condition mismatch unless the encoding is checked against an external source. Similarly, failure to retrieve a rule cannot be distinguished from the absence of a usable rule solely from an empty admitted set; broader rule search or a complete check of the relevant library is required. The diagnostic retest in Section 6.4 widens candidate coverage; it is not a complete oracle over all possible rules and bindings.

The practical advantage of this representation is that failure becomes an inspectable state and control question: which premise was needed, which candidate was considered, which applicability check rejected it, and how far the search progressed. Those observations guide knowledge completion, rebinding, search reorganization, or capability-library updates, and each is reported together with the cause it can establish.

### 4.4 What dynamic hierarchy changes in search

ERS uses high-level control to identify relevant subgoals, restrict or reorder candidate sets, and redirect exploration when execution feedback reveals a new premise, an invalid branch, or a changed goal. Its advantage is the organization of a combinatorial search rather than the elimination of combinatorial growth.

Compare ERS with a flat procedure that searches the same admissible rules without building a goal-dependency hierarchy. Let $B_t$ be an upper bound on candidates expanded at depth $t$ in a flat procedure and $K_t$ an upper bound on those prioritized by control in a comparable expansion phase. If a valid route remains covered and $K_t < B_t$, the corresponding *candidate-sequence upper bounds* are

$$N_{flat} \leq \prod_{t=1}^{L} B_t, \qquad N_{ctrl} \leq \prod_{t=1}^{L} K_t. \tag{18}$$

These inequalities bound candidate counts rather than runtime, and whether the two algorithms actually visit their upper bounds is a separate question. A hierarchical choice has an overhead: decomposing the goal, checking control applicability, evaluating candidates, and sometimes calling an additional model. For a transparent simplified comparison, let $c_{test}$ be a common cost of expanding/checking a candidate, and let $c_{control}$ be the entire additional control overhead for the phase. If the flat procedure checks $B_t$ candidates and hierarchical control checks $K_t$, with equal per-candidate cost otherwise, hierarchical control is cheaper **under this cost model** exactly when

$$c_{control} < (B_t - K_t)c_{test}. \tag{19}$$

The condition expresses the engineering tradeoff: a smaller candidate set is useful when the avoided expansion cost exceeds the overhead of organizing it. Whether hierarchy pays therefore depends on the relation between control overhead and avoided expansion cost, which (19) makes explicit. For a tree search with a branching bound $\beta > 1$ and lookahead horizon $h$, the number of candidate expansions through depth $h$ obeys

$$N_h \leq \sum_{d=1}^{h} \beta^d = O(\beta^h). \tag{20}$$

Deeper lookahead can therefore reduce unnecessary *executed domain operations* while increasing search effort; Section 6.2 measures domain steps at several horizons, not total computation. The cost of search also depends on repeated candidate retrieval, lookahead expansion, and evidence checks, even when two control configurations complete the same domain computations. Retaining valid intermediate state and making decisions within the relevant subgoal can reduce these repeated evaluations, but any savings must be weighed against the additional control operations. The architectural significance of dynamic control is that it can direct a finite budget toward currently relevant dependencies, adjust the hierarchy through feedback, and assemble new paths from unchanged basic capabilities. The compositional and revision checks in Sections 6.6–6.7 connect the open path space of Section 4.1 with the state-maintenance mechanisms of Sections 3.2 and 3.5. Section 6.8 measures the search-organization tradeoff in inference time and operation counts on development tasks; Section 7.1 describes the remaining budget and end-to-end evaluation scope.

## 5 Experimental Setup

### 5.1 Tasks and evaluation protocol

We design the original batches to address whether local rules form long and cross-subgoal paths, how lookahead changes executed paths, and whether knowledge completion respects evidence constraints. Two focused checks examine new input/path combinations under a frozen rule network and the handling of changed or withdrawn facts. These checks belong to a broader 30-run engineering audit, supplied with the supplementary material, and each reported group retains its own denominator rather than being pooled with the earlier experiments.

We use author-constructed structured event graphs, domain rules, and property definitions throughout. Scoring answers and reference trajectories are excluded from inference, and evaluation checks goals, premises, bindings, and executed paths. The front-end language model and automatic rule learning are disabled, while backend matching uses a local vector model. Because the graphs are constructed rather than extracted, the results locate behavior in the reasoning backend and not in the input interface, where an extracted binding could be wrong in a way no admission check would detect; the reported measurements do not cover that case.

Long-chain tasks require 2, 4, 8, or 12 dependency steps, with five graph variants each. Medium-scale compositions span inventory, supply-chain, finance, and cross-domain tasks, and comprise 12 calibration cases, 30 positive cases, and 30 negative cases. The graphs share topologies, so variation is structural rather than across independent task families. The focused additional checks use constructed backend inputs and frozen rules rather than newly learned operators or an end-to-end language front end.

## 5.2 Execution configuration

**Table 2.** Execution configuration by batch. Scheduling rounds and candidate-rule checks are different units and are not comparable across rows.

| Experiment / setting | Parameter | Value | Unit / scope |
|---|---|---|---|
| Default reference | Scheduling budget | 128 | rounds (reference) |
| Long-chain | Scheduling budget | 384 | rounds |
| Medium composition | Scheduling budget | 512 | rounds |
| Lookahead comparison | Scheduling budget | 512 | rounds (shared) |
| Lookahead comparison | Horizon | 0, 8, 16 | lookahead steps |
| Stress batch | Candidate-rule checks | 8 | per completion |
| Candidate-range retest | Candidate-rule checks | 64 | per completion |
| Additional focused checks | Loop ceiling | 192 | iterations per run |
| Additional focused checks | Lookahead horizon | 8 | steps |
| Development control comparison (development_r4_typed) | Scheduling budget / lookahead | Not reported in the available audit | Not reported in the available audit |
| State-change control comparison (intervention_r1) | Scheduling budget / lookahead | Not reported in the available audit | Not reported in the available audit |

*For the experiments in Sections 6.1–6.7, target information reaches local selection through the recorded state and rule conditions. A data-propagation defect identified during the original audit was corrected before those runs, and the results in those sections use the corrected route. The separate frozen implementations in Section 5.5 are evaluated from their own recorded runs.*

## 5.3 Evaluation metrics

Answer-and-evidence acceptance checks the result and its support; strict mechanism acceptance additionally requires the specified premises and operations in the executed path. Let $\alpha_i$ indicate answer-and-evidence acceptance for case $i$ and $\mu_i$ indicate acceptance under predefined mechanism criteria; then

$$A_{acc} = \frac{1}{N}\sum_{i=1}^{N}\alpha_i, \qquad M_{mech} = \frac{1}{N}\sum_{i=1}^{N}\mu_i. \tag{21}$$

Positive cases require a correct, supported result; negative cases require withholding unsupported target outputs. We report the two criteria separately. Because strict mechanism acceptance additionally requires answer-and-evidence acceptance, $\mu_i \leq \alpha_i$ by definition. Hence $A_{\mathrm{acc}} - M_{\mathrm{mech}} = \frac{1}{N}\sum_{i=1}^{N}(\alpha_i - \mu_i)$ measures the proportion of cases that satisfy answer-and-evidence acceptance but fail the prespecified mechanism criterion; their count is $N(A_{\mathrm{acc}} - M_{\mathrm{mech}})$.

In the 48-case stress batch, the observed gap is $2/48$, corresponding to two cases. Such a difference may reflect an invalid process or an admissible path that fails an exact-trajectory requirement; the execution trace must be inspected to distinguish them (Section 6.5).

The case classes contribute unequally: a system that withholds on every case scores 24/24 on the negatives and 0/24 on the positives, obtaining $A_{acc} = 0.50$ without solving anything, and that value is the trivial-abstention baseline against which the reported figures must be read. Abstention on the negatives and low completion on the positives also share a cause, since a stricter admission predicate reduces false assertions and successful completions together. We therefore report the positive completion and unsupported-negative-output counts alongside the combined acceptance rate, and give their per-class decomposition in Table 5.

Path efficiency is measured by executed domain steps and by irrelevant steps, defined as steps on branches that no accepted result depends on. Relative step reduction is

$$R_{step} = \frac{S_{base} - S_{config}}{S_{base}}, \qquad S_{base} > 0. \tag{22}$$

This metric counts domain steps only. Candidate retrieval, lookahead evaluation, and model calls are excluded from it, and the resulting scope is clarified in Section 7.

## 5.4 Focused checks of recombination and state revision

We report 18 targeted cases within a 30-run engineering review: twelve frozen-rule checks comprising nine accepted derivations involving unseen input/path combinations and three justified refusals, together with six scenarios involving changed, withdrawn, or newly supplied premises. The other twelve records belong to the same review and concern control configurations; they are distinct from the twelve intervention_r1 records and the development comparisons described in Section 5.5. The focused results are reported with their own denominators.

For the eighteen targeted checks, we held the rule library and formal database fixed. The read-only audit examined semantic feature identities used in bindings, recursive source validity, active computed values, and required-condition evidence. No frontend model, training step, or persistent rule update was used. The recombination and fact-change cases test different mechanisms and are not pooled into a single success rate.

### 5.5 Dynamic control comparison

We evaluate three control configurations on the same structured development-task inputs. DYN uses dynamic subgoal control with local candidate decisions, retained state, and feedback-guided adjustment; GLOB uses a single-level global goal-directed controller; and STAT selects executable rules by strict static priority without high-level goal-directed control. The STAT configuration can still execute multiple low-level rules, so it is not a single-step baseline. These settings compare control organization within the implemented backend rather than the performance of all possible flat or static algorithms.

The development_r4_typed batch contains sixteen tasks run under each of DYN, GLOB, and STAT, for 48 recorded runs. The paired cost analysis uses the fourteen tasks that both DYN and GLOB solve correctly through all required stages. Two other development tasks are excluded from solved-cost accounting: the supplied audit excludes information-insufficient and budget-exhausted cases, but does not give separate sixteen-task completion totals for each configuration. The STAT workload is recorded on the same fourteen inputs, without a separately established STAT success denominator.

The independent intervention_r1 batch contains four execution-time-change tasks run under all three configurations, for twelve additional records. These two batches use different frozen source versions and are reported separately, rather than as repeated trials or as part of the 30-run focused audit in Section 5.4. The supplied cost analysis does not specify the scheduling budgets or lookahead settings of the two frozen implementations. Their results are therefore comparisons under the recorded configurations, not a verified matched-budget comparison.

We distinguish inference wall-clock time and process CPU time from whole-worker time, which also includes process startup, loading, recording, and exit. Search-activity counts include candidate-catalog checks, lookahead-node counts, strict candidate checks, and evidence-validity calls. Domain failures count recorded unsuccessful domain-execution attempts, not task-level failures or candidates rejected before execution. We also report correct domain computations and control actions. The paired-success comparison describes cost *conditional on both configurations solving the task*; it does not establish their completion rates over all sixteen tasks. These measurements cover the frozen structured-event backend configurations, not the separate public inference demo or the cost of an end-to-end semantic-input pipeline.

## 6 Results

### 6.1 Long-chain and compositional reasoning

**Table 3.** Long-chain and composition outcomes. The six financial cases are a subset of the medium-scale batch and are not an additional batch.

| Experiment | Task composition | Result |
|---|---|---|
| General control | 5 groups × 5 cases | 25/25 accepted |
| Long-chain | 2, 4, 8, 12 steps × 5 variants | 20/20 accepted |
| Small-scale composition | 4 calibration / 7 positive / 5 negative | 16/16 accepted |
| Medium-scale composition | 12 calibration / 30 positive / 30 negative | 72/72 accepted |
| Financial subset | 6 positive DSO and cash-conversion cases | 6/6 accepted on value, unit, provenance |

We observe 4–14 domain operations in medium-scale positive paths, and the six financial cases also pass value, unit, and provenance checks. Because calibration and negative cases are included in the totals, the 72 accepted cases do not represent 72 new positive compositions. At the 12-step setting, execution used 15 domain operations, three of which lay on a branch that no accepted result depended on, so a completed path was not always an efficient one. Our author-constructed graphs demonstrate feasible rule composition; independent path and topology holdouts would be needed to test generalization.

### 6.2 Control strategy variants

Lookahead horizons of 0, 8, and 16 satisfy all ten task-acceptance checks. Table 4 compares domain steps over the eight solvable positive tasks.

**Table 4.** Lookahead comparison on the same ten tasks. Step counts are over the eight solvable positives.

| Lookahead | Accepted (10 cases) | Domain steps (8 positives) | Irrelevant steps | Reduction vs. previous setting | Reduction vs. horizon 0 |
|---|---|---|---|---|---|
| 0 (basic control) | 10/10 | 94 | 32 | — | — |
| 8 steps | 10/10 | 64 | 8 | 31.9% | 31.9% |
| 16 steps | 10/10 | 56 | 0 | 12.5% | 40.4% |

Using (22), domain steps fall by 31.9% from 94 to 64 between horizons 0 and 8, and by a further 12.5% to 56 between horizons 8 and 16. We take horizon 0 as basic control. The irrelevant-operation column shows zero such operations at horizon 16 in these recorded trajectories, and each counted operation is classified by whether it supports an accepted result. The domain-step percentages, by contrast, are measured in a unit that excludes candidate search, lookahead evaluation, and model calls; Section 7.1 takes up the total-cost comparison.

Figure 5 schematically contrasts immediate-state expansion with lookahead-guided evaluation and pruning. It illustrates the selection mechanism behind the domain-step measurements in Table 4; the figure is not an additional measurement of candidate evaluations or total runtime.

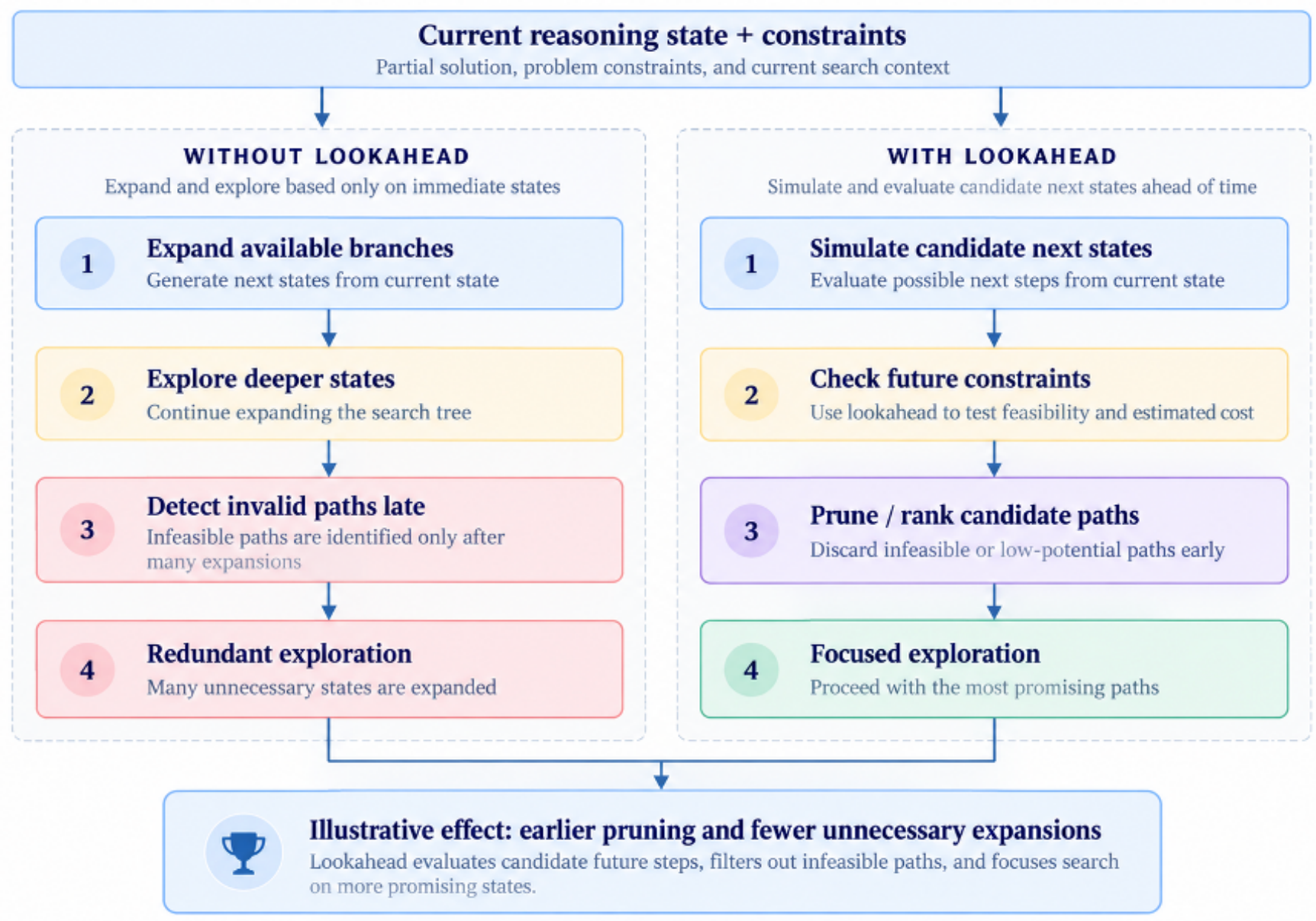

**Figure 5.** Conceptual effect of prospective (lookahead) search. Without lookahead, an unsuitable branch may be explored before its limitations become apparent. With lookahead, candidate future states can be checked and ranked before committing to domain execution, allowing the controller to prioritize paths with stronger expected goal progress. The measured effect on executed steps at horizons 0, 8, and 16 is reported in Table 4.

### 6.3 Knowledge-assisted reasoning

We build the stress cases around information gaps, cross-entity and cross-period interference, multi-round knowledge completion, and the reuse of historical paths. The 48 cases comprise six groups of eight: 24 solvable positives and 24 negatives. Answer-and-evidence acceptance is 42/48, strict mechanism acceptance is 40/48, and positive completion is 18/24; none of the 24 negatives produces an unsupported target output. Four records from successful executions form frozen memory, retaining the historical fact identities.

**Table 5.** Knowledge-and-memory stress outcomes, with the per-class decomposition. The final row gives the trivial-abstention baseline that the reported figures must be read against.

| Metric | Result | Scope |
|---|---|---|
| Answer-and-evidence acceptance $A_{acc}$ | 42/48 | supported task outputs |
| Strict mechanism acceptance $M_{mech}$ | 40/48 | prespecified operations |
| Difference $A_{acc} - M_{mech}$ | 2/48 | accepted outcome; prespecified mechanism unmet |
| Positives: asserted / withheld | 18 / 6 | of 24 solvable positives |
| Negatives: asserted / withheld | 0 / 24 | of 24 negative cases |
| Trivial-abstention baseline on this batch | 0.50 | withholds every case |

The two acceptance figures decompose exactly. Of the 42 cases meeting the answer criterion, 24 are negatives on which output was correctly withheld and 18 are positives; of the 40 meeting the mechanism criterion, the same 24 negatives are counted and 16 positives remain. The difference of two cases therefore identifies positives whose answer was supported but whose executed path did not match the prespecified operation sequence.

Two readings follow, and both are needed. The first is the completion figure: the system withheld on six positives while asserting nothing unsupported on the 24 negatives, at a completion rate of 75%. Reporting both outcomes avoids mistaking universal abstention for task-solving ability. The second is that a difference of two cases is an existence observation rather than a statistical estimate: the batch illustrates the distinction rather than estimating its frequency in a broader population. What the difference establishes is that answer-level and mechanism-level success are distinct classes and both non-empty, which is what the two-criterion design was introduced to detect.

### 6.4 Failure attribution

The six positives missed in Section 6.3 were missed under a limit of eight candidate-rule checks per knowledge-completion operation. A targeted retest raises the limit to 64 with all other settings unchanged, and includes the six previously missed cases together with six matched negatives.

**Table 6.** Candidate-range diagnostic on 12 retested cases, separate from the stress batch.

| Check | Result | Scope |
|---|---|---|
| Task acceptance | 12/12 | 6 positives + 6 negatives |
| Solvable-positive completion | 6/6 | previously missed cases |
| Unsupported negative outputs | 0/6 | matched negative cases |
| Two knowledge-completion rounds | 6/6 | positive trajectories |

*The read-only audit performed 28 completion-proof checks and 52 slot-binding checks across the positive trajectories.*

All twelve cases succeed under the raised limit, which locates the original failures in candidate coverage rather than in absent local rules: the rules required to solve the six positives were present in $R_{exec}$ throughout, but $D_t$ did not retrieve them at the original limit. We read the raised limit in the terms of Section 4.3: it is a targeted candidate-range diagnostic rather than an exhaustive retrieval oracle, and it tests this possible failure cause on the observed cases. Two limits belong with the result. The expanded search carries an additional cost that we do not quantify here, and the twelve cases were selected by the failure they are used to explain, so they are not independent test samples.

### 6.5 Mechanism, outcome, and the choice of scoring criterion

Task success, path validity, and the prespecified mechanism checks are three distinct quantities, and the stress batch contains a case that separates them: ERS completed a valid path that had been recorded historically and in the same episode also executed other operations that were admissible at the time, so an exact-trajectory scorer rejected the mechanism criterion even though the answer was supported. A second case constrains the interpretation, in that an alternative path was taken before rather than after a calculation error, so post-error recovery is not established by this batch.

The comparison with supervised fine-tuning concerns evaluation rather than efficiency. Fine-tuning on fixed demonstration trajectories treats one admissible path as the supervisory target [21], whereas ERS explores operations that satisfy the current rule conditions and records their dependencies; the first case above shows what happens when such exploration is scored against a single reference. If admissible paths beyond the demonstration are to be discovered at all, the scoring rule has to check support rather than resemblance. Our experiments measure the difference between these two criteria; efficiency and generalization relative to fine-tuning lie outside their scope.

### 6.6 Unseen input and path combinations with frozen rules

In the frozen-rule check, all twelve cases meet their case-specific acceptance criteria: nine accepted derivations and three justified refusals (blind_10, blind_11, and blind_12).

For cases requiring an accepted answer, the complete rule–binding path is evaluated against $\Pi_{\mathrm{stored}}$, and its rule sequence against the previously instantiated sequences in $\mathcal{L}_{\mathrm{seen}}$, as distinguished from the control-permitted space $\mathcal{L}_{\mathrm{ctrl}}$ in Section 4.1.

The negative controls instead require an explicitly justified refusal: their control traces record a blocking reason, and reaching a scheduler exit or loop cap alone does not count as valid refusal. The supplemental audit checks the actual semantic feature identities used in slot bindings, rather than merely the existence of referenced IDs. The twelve-case denominator therefore covers both accepted derivations and justified refusals under the stated criteria.

Earlier work measures compositional generalization as transfer to novel combinations of primitives seen separately [16, 17, 18]; that is a property of a learner. We scope the present result to new compositions and admissibility checks within a fixed, finite rule network, which concern the construction of valid paths and the rejection of unsupported ones.

The accepted derivations demonstrate the assembly of previously unstored routes from reusable local rules, while the negative controls test whether the same system respects the conditions that limit those compositions.

### 6.7 Fact-change propagation and active-state validity

Six targeted scenarios exercise how changes to the premises of a derived result affect its current acceptance. The audit checks the updated answer, recursive source validity, and consistency of active computed values in every case.

**Table 7.** Six fact-change checks, with verdicts from the additional read-only audit. All six pass the three checks on updated results, recursive provenance, and active computed values.

| Scenario | Intervention | Audit result |
|---|---|---|
| dynamic_withdraw_premise | Withdraw a prerequisite fact | PASS |
| dynamic_correct_number | Correct an upstream numerical value | PASS |
| dynamic_irrelevant_change | Change a fact irrelevant to the target support | PASS |
| dynamic_independent_support | Retain an alternative independent justification | PASS |
| dynamic_revoke_approval | Revoke a required approval condition | PASS |
| dynamic_late_premise | Supply a previously missing premise | PASS |

The audit finds **6/6** correct outcomes under the specified checks; no required-condition evidence is missing from the reported computation-source links. We connect this result to the explicit support semantics of Section 3.2 and the dependency-scoped revision principle of Section 4.2. A source identifier alone does not prove that its old value remains valid, so the review checks active values and premises in addition to provenance links. External snapshot retention in the review should not be interpreted as confirmation of native long-term historical-state management inside ERS. These cases establish the reported backend fact-change behavior.

### 6.8 Dynamic control and search efficiency

We compare the measured work of the dynamic subgoal and single-level global control configurations on identical structured development inputs. On the fourteen tasks solved correctly by both DYN and GLOB in all stages, each performs 243 correct domain computations. Table 8 reports accumulated costs across these tasks; the percentages are ratios of the totals, not medians of per-task ratios. Scheduling budgets and lookahead settings have not been recovered from the frozen configurations, so the observed differences are not identified as a controlled, matched-budget effect.

**Table 8.** Recorded costs for fourteen jointly solved structured development tasks under their respective frozen configurations. DYN denotes dynamic subgoal control and GLOB denotes single-level global control. Counts are accumulated across these fourteen tasks rather than all sixteen; negative changes indicate a lower recorded quantity for DYN.

| Measure | DYN | GLOB | DYN relative to GLOB |
|---|---|---|---|
| Inference wall-clock time | 49.008 s | 101.215 s | −51.6% |
| Inference process CPU time | 48.454 s | 101.101 s | −52.1% |
| Whole-worker time | 124.229 s | 174.376 s | −28.8% |
| Candidate-catalog checks | 874 | 3,555 | −75.4% |
| Lookahead-node counts | 218 | 3,342 | −93.5% |
| Strict candidate checks | 533 | 723 | −26.3% |
| Evidence-validity calls | 4,910 | 13,844 | −64.5% |
| Control actions | 389 | 279 | +39.4% |
| Correct domain computations | 243 | 243 | unchanged |
| Domain failures | 4 | 8 | −50.0% |

DYN records more control actions but fewer candidate-catalog checks, lookahead-node evaluations, and evidence-validity calls. Within the recorded phase timers, control and evaluation account for 35.073 s versus 87.641 s, while domain execution accounts for 11.981 s versus 11.759 s (DYN versus GLOB). The residuals between those two phase totals and inference wall-clock time are 1.954 s and 1.815 s, respectively; they are unallocated recorded time, not a separately measured operation. The observed time differences coincide with less search and evaluation activity without isolating a causal contribution from any single control mechanism.

On the same fourteen input tasks, STAT records 1,508.053 s of inference wall-clock time and 2,301 correct domain computations, with substantial work on non-target branches. These are descriptive workload totals for the implemented static-priority selector. Because the supplied aggregate does not establish how many of the fourteen tasks STAT solved, its totals cannot be used as a like-for-like per-success cost comparison with DYN or GLOB.

All three configurations complete the four execution-time-change tasks correctly in the separate intervention_r1 batch. Accumulated inference wall-clock time is 11.914 s for DYN and 20.213 s for GLOB, an observed 41.1% difference. These records come from another frozen implementation and are not pooled with Table 8. In one intervention DYN executes more domain operations than GLOB, so lower recorded time does not imply the shortest execution path or a uniform cost advantage.

## 7 Limitations and Roadmap

We draw three directions for extending the present account of compositional reasoning from the structured-event results. These directions follow from the specific mechanisms formalized in Sections 3–4 and the targeted findings in Section 6.

**7.1 Search organization and resource allocation.** In Section 6.2, lookahead reduces executed domain operations by 40.4% between horizons 0 and 16 on eight positive tasks, while irrelevant operations fall from 32 to zero in the evaluated batch. Section 6.8 reports a separate development comparison: on fourteen tasks solved by both configurations, DYN records lower candidate-evaluation counts and inference time than GLOB while both perform 243 correct domain computations. These differences are measured under the respective recorded frozen configurations; scheduling and lookahead budgets have not yet been verified as matched, so they should not be attributed solely to control organization. The tradeoff in (19) remains dependent on task structure, budget, and control overhead. Matched-budget comparisons, independent unseen topologies, repeated trials, the STAT completion denominator, and full semantic-input-to-answer cost remain to be evaluated.

**7.2 Compositional coverage.** The frozen-rule evaluation in Section 6.6 comprises nine accepted derivations and three justified refusals, with all twelve cases meeting their respective criteria. The intended next step is to characterize how coverage changes across larger dependency structures, independent task families, and compositional splits. We intend to distinguish the construction of new *paths* from the induction of new *rules*, which are separate capabilities in ERS. The reported cases are author-constructed, so the next measurement we plan is an independent holdout that separates task families by topology, with the acceptance criteria fixed before evaluation.

**7.3 Dependency-preserving revision.** The 6/6 fact-change checks in Section 6.7 cover premise withdrawal, correction, irrelevant updates, independent support, approval revocation, and late-arriving evidence. They demonstrate the specified active-state and provenance checks for these scenarios. We plan next to measure the affected dependency closure during revision and to compare its resource cost with full recomputation. Historical-path memory can likewise be examined separately as a mechanism for prioritizing future rule combinations.

Together, these directions retain the central architectural question: how finite local capabilities, explicit evidence dependencies, and task-dependent control can support new solutions and maintain their validity as the problem changes.

## 8 Conclusion

Complex reasoning can be constructed from finite, reusable local capabilities through dynamic hierarchical composition over explicit event states. We take this as the central claim of ERS. Low-level rules specify admissible changes to the current facts; high-level control rules determine which subgoals and rule combinations matter for the present objective. The hierarchy grows and reorganizes with the problem, and neither the number of levels nor the resulting solution path must be fixed in advance.

We treat the reasoning process as an executable, inspectable dependency structure: Semantic representations guide candidate matching, while explicit conditions determine execution admissibility. Accepted results preserve their evidential dependencies and support local revalidation when facts change. Only affected justifications need to be reconsidered, while conclusions with independent valid support remain accepted. Through top-down path organization and bottom-up execution feedback, ERS enables dynamic, continuous reasoning rather than the reuse of pre-stored solutions.

The reported backend results include long legal rule chains, 12/12 successful frozen-rule checks comprising nine accepted derivations involving unseen input/path combinations and three justified refusals, and 6/6 passed fact-change checks for updated values and their supporting evidence. These targeted observations show that the same finite set of local capabilities can be recomposed into previously unstored task paths, that unsupported requests can be rejected under explicit conditions, and that conclusions can be revalidated against changing premises. In the paired development-task comparison, the recorded dynamic subgoal configuration exhibits lower candidate-evaluation counts and inference time than the single-level global configuration under their respective frozen settings, with the same number of correct domain computations. ERS thereby takes its place alongside language-based reasoning, symbolic planning, and learned predictive world models as a concrete architecture whose reasoning paths are composed from finite reusable rules and maintained by task-dependent control.

### Author Contributions

**Cheng Lyu** originated the central research idea, led the conceptualization and theoretical development of the ERS architecture, designed and conducted the experiments, analyzed and interpreted the experimental results, and led the writing and revision of the manuscript.

**Xueqi Wang** contributed to discussions of some aspects of the model's conceptual framework and participated in the writing and revision of the manuscript.

## Appendix A Illustrative Example of Hierarchical Event Reasoning

This appendix illustrates how ERS constructs and validates a reasoning path through hierarchical rule composition and explicit state updates. A financial calculation task demonstrates goal decomposition, rule binding, evidence constraints, and dependency preservation. The example illustrates the reasoning mechanism and is separate from the results reported in Section 6.

### A.1 Task definition and initial event state

We use a task requiring the calculation of Company A's days sales outstanding (DSO) for fiscal year 2025. The available information consists of beginning accounts receivable of 40, ending accounts receivable of 60, FY2025 revenue of 500, and FY2024 revenue of 450, all expressed in units of ten thousand currency units. The task explicitly authorizes endpoint averaging and a 365-day conversion convention. The target calculation is

$$DSO = \frac{\text{average accounts receivable}}{\text{FY2025 revenue}} \times 365.$$

We represent the input as an event state containing numerical values, entity identities, reporting periods, units, and evidence qualifiers. The objective is to produce a DSO value satisfying the specified calculation conventions and temporal constraints. The FY2024 revenue is semantically relevant to the financial domain but does not satisfy the task's required period, and its presence is the point of the example: similarity alone cannot determine whether a piece of information may participate in a valid reasoning step.

### A.2 Hierarchical rule composition and execution

We organize the objective through reusable high-level and low-level rules instead of requiring an independently stored solution for the complete DSO task. High-level control identifies three dependent subgoals: obtaining average accounts receivable, computing the receivables-to-revenue ratio, and converting the ratio into days. Each subgoal is associated with a low-level rule whose execution requires valid object bindings, applicable evidence, and satisfaction of its declared conditions. Figure A1 shows the dependency structure and the propagation of accepted intermediate results through the event state.

**Table A1.** Rule execution, validation, and state updates in the illustrative DSO task.

| Stage | Operation | Validation and state update |
|---|---|---|
| Goal organization | High-level control | Decompose DSO into averaging, ratio, and day-conversion subgoals |
| Input binding | Evidence and task constraints | Admit FY2025 revenue; exclude FY2024 revenue |
| Step 1 | Average receivables | $(40 + 60)/2 = 50$; record input dependencies |
| Step 2 | Calculate ratio | $50/500 = 0.1$; verify entity, period, and units |
| Step 3 | Convert to days | $0.1 \times 365 = 36.5$ days; preserve calculation dependencies |
| Acceptance | Goal and evidence verification | Accept 36.5 days under the specified task constraints |

The process is a sequence of accepted state transformations, $S_0 \rightarrow S_1 \rightarrow S_2 \rightarrow S_3$. Here $S_0$ contains the initial facts and task constraints; $S_1$ additionally contains the verified average receivables of 50; $S_2$ contains the verified ratio of 0.1; and $S_3$ contains the final DSO value of 36.5 days. We perform each transition with an applicable low-level rule that records its actual supporting premises, so the output of one accepted operation becomes an available input for subsequent operations while high-level control maintains the dependency order.

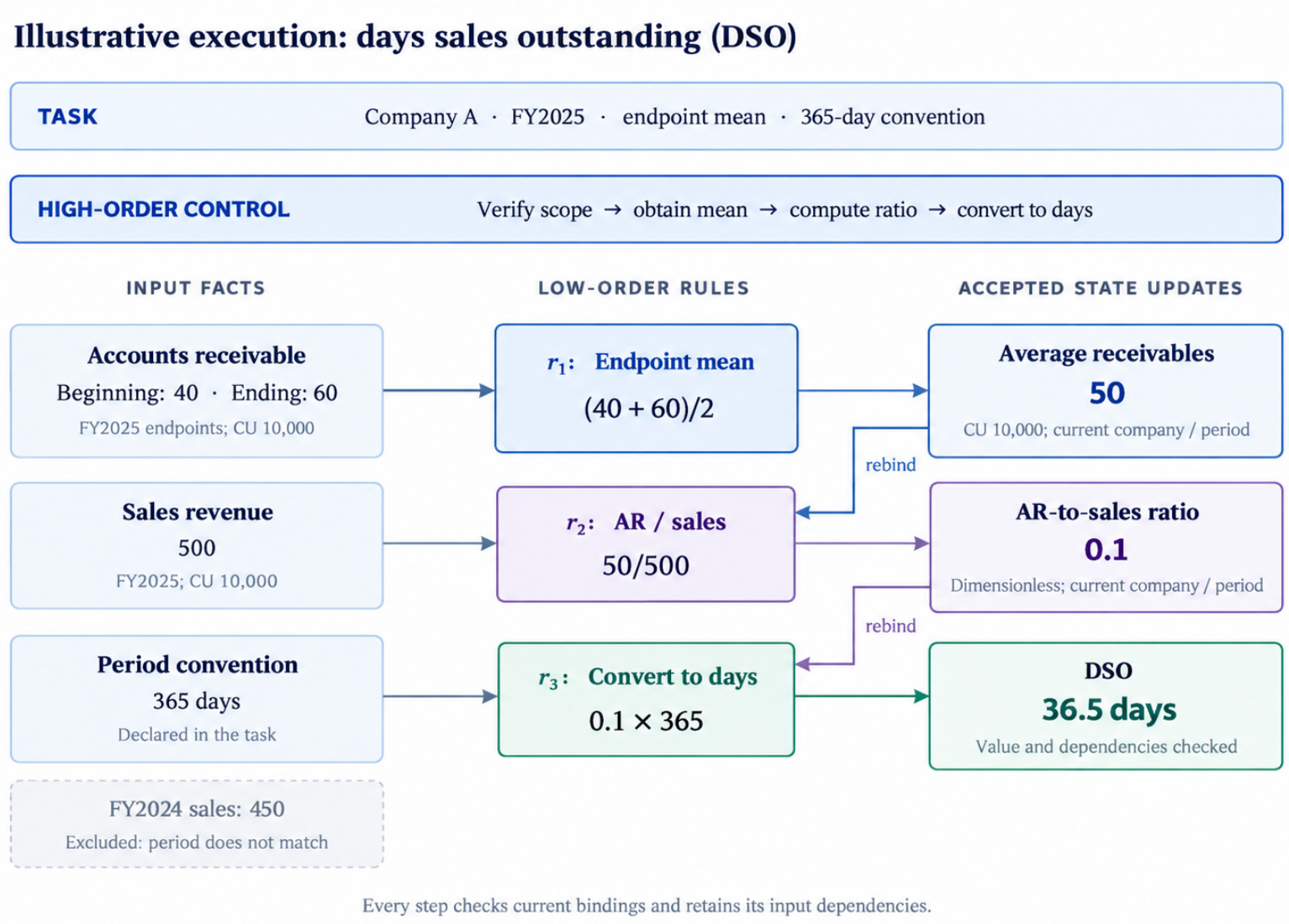

**Figure A1.** Hierarchical rule execution and dependency propagation in the DSO example. High-level control organizes the target into three dependent subgoals. Low-level rules perform averaging, ratio calculation, and conversion to days. Each accepted output becomes an available premise for subsequent execution, while entity, period, and evidence constraints are preserved.

### A.3 Constraint validation and dependency preservation

Three properties are visible in the trace. First, semantic relevance does not imply execution eligibility: both revenue figures relate to the same financial concept, and only one satisfies the required period, so the FY2024 value is excluded at the binding check before it can enter any accepted path.

Second, intermediate results stay connected to their supporting facts and operations—the average receivables to the two authorized observations, the ratio to the verified average and the FY2025 revenue, the final DSO to the accepted ratio and the day convention—which makes the path inspectable and distinguishes a valid numerical result from one obtained through incompatible inputs.

Third, the dependency structure supports targeted revalidation: were the FY2025 revenue corrected, the ratio and the DSO would require revalidation while the average receivables would remain valid if its observations were unchanged, as described by the dependency-scoped revision property in Section 4.2. We take the significance of the example to lie in the construction and verification of the process rather than in the numerical calculation alone.

## Appendix B ERS in Relation to World Models

Section 2.3 summarizes three axes of comparison; this appendix records the full set for readers working primarily in the world-model literature. As in Section 2.3, the comparison is architectural: each entry follows from the stated design assumptions of the family in question, and none of the entries is a measured result here.

**B.1 Reasoning mechanism and logical reliability.** World models learn abstract representations of environments and their dynamics, enabling future-state prediction in latent space [22, 25]. Implementations of this kind rely on learned predictive functions for implicit state transitions, where statistical plausibility does not guarantee the logical validity of an individual step or of its evidential dependencies. ERS composes reasoning from executable rules, so each operation is checked against its premises, bindings, and resulting dependencies.

**B.2 Hierarchical organization and dynamic problem solving.** Hierarchical proposals in this family rely on fixed levels or learn a hierarchy end to end [23, 27], and their latent representations give limited support for identifying a failed premise, a conflicting rule, or a change in logical dependency at the level of individual operations. ERS builds its hierarchy at run time from the task's dependency structure, so an unmet premise or a new information gap triggers revision of the relevant subgoals rather than merely degrading performance.

**B.3 Capability expansion and long-term maintenance.** World models that rely on parameterized representations may need additional training or fine-tuning to incorporate new domain knowledge, and such updates can interfere with previously acquired behavior. ERS organizes knowledge, execution rules, and control strategies as independently manageable components, so domain capability is extended by adding, modifying, or replacing a module rather than by retraining the semantic model. Modules no longer required can be unloaded, which changes the candidate range and the search cost with it.

**B.4 Integration of semantic understanding and logical execution.** Neural models extract continuous semantic features from open-ended input, while symbolic systems supply explicit operational conditions, and each is limited in what it can check or in the flexibility of its input. ERS connects them through a unified event representation: encoders identify objects, attributes, and relations, while the backend selects, executes, and validates rules against the current state. Similarity discovers candidates and eligibility is decided by bindings, applicability, and evidence, so the two mechanisms are maintained independently.

**B.5 Error location and corrective response.** In systems driven by latent state evolution, the internal cause of a failed prediction is difficult to read from the representation. We separate factual errors, missing knowledge or rules, unsatisfied execution conditions, and failure to discover a path by recording object bindings, rule applicability, execution results, and cross-step dependencies. These map to different responses—completion, capability extension, renewed matching, or a change of control strategy—which is the mechanism behind the attribution of Sections 4.3 and 6.4.

**B.6 Dynamic state maintenance.** Facts and constraints change over time, and a system without explicit dependency records cannot readily determine which derived conclusions are affected. ERS maintains object identities, validity intervals, provenance, and execution dependencies in the event graph, so a modified, withdrawn, or invalidated fact leads to reassessment of the dependent results while conclusions with an independent valid support path remain accepted. This avoids both unnecessary recomputation and the incorrect assertion that a withdrawal is a refutation.

**B.7 Evaluation of reasoning and acceptance of output.** Generative systems are often evaluated by answer matching, reference trajectories, or external scorers, yet a correct answer does not imply a valid process and a solution differing from a reference trajectory is not thereby incorrect. Explicit rule execution and dependency records let ERS examine goal satisfaction, local operation validity, and evidential support separately, accepting an alternative path that satisfies the constraints while rejecting a conclusion that lacks necessary premises.

**B.8 Uncertainty and information gaps.** Real-world tasks frequently involve incomplete information, conflicting evidence, and unsatisfied prerequisites. ERS distinguishes confirmed facts, hypotheses, unresolved relations, and supported conclusions in the event graph, and evidence status and absent premises bear directly on rule eligibility and strategy selection rather than being attached as a confidence estimate after generation. When support is insufficient the system can block unsupported operations, attempt completion, reorganize subgoals, or retain an undetermined status, and a blocked path can be reconsidered once valid information arrives.

**Complementary roles.** Learned world models can supply structured or latent predictions about possible environmental states, and ERS can use supported predictions as candidate information subject to explicit source and applicability conditions. The architectural contrast concerns how a proposed state transition becomes an accepted inference: prediction and plausibility can guide exploration, while an ERS rule path records the premises, control choices, and dependencies under which that inference is licensed. The two directions can therefore inform one another without being treated as identical tasks or directly interchangeable models.